\documentclass[final]{siamltex}
\usepackage{graphicx}
\usepackage{color}
\usepackage{amssymb,amsmath}
\usepackage[linesnumbered,boxed]{algorithm2e}
\usepackage{subfigure}
\usepackage{multirow}

\newcommand{\F}{{\cal F}}
\newcommand{\cS}{{\cal G}}
\newcommand{\T}{{\cal T}}

\title{Geometric Tight Frame based Stylometry for Art Authentication of van Gogh Paintings}

\author{Haixia Liu\thanks{Department of Mathematics, The Chinese University of Hong Kong, Hong Kong, China.
 Research of Raymond H. Chan is supported in part by HKRGC GRF Grant No. CUHK400412, HKRGC CRF Grant No. CUHK2/CRF/11G,
HKRGC AoE Grant AoE/M-05/12, CUHK DAG No. 4053007, and CUHK FIS Grant No. 1902036. ({\tt mahxliu@ust.hk, rchan@math.cuhk.edu.hk}).}
\and Raymond H. Chan\footnotemark[1]
\and Yuan Yao\thanks{School of Mathematical Sciences, LMAM-LMEQF-LMPR, Peking University, Beijing, China 100871. The research of Yuan Yao is supported in part by National Basic Research Program of China (973 Program 2012CB825501,2015CB856000), NSFC grant 61071157 and 61370004. ({\tt yuany@math.pku.edu.cn}).}}

\begin{document}

\maketitle

\begin{abstract}
{This paper is about authenticating genuine van Gogh paintings from forgeries. The authentication
process depends on two key steps: feature extraction and
outlier detection. In this paper, a geometric tight frame and some simple statistics of the tight frame
coefficients are used to extract features from the paintings.
Then a forward stage-wise rank boosting is used to select
a small set of features for more accurate classification so
that van Gogh paintings are highly concentrated
towards some center point while forgeries are spread out as outliers.
Numerical results show that our method can achieve 86.08\% classification accuracy under the leave-one-out cross-validation procedure. Our method also identifies five features that are
much more predominant than other features.
Using just these five features for classification, our method can give 88.61\% classification accuracy
 which is the highest so far reported in literature.  Evaluation of the five features is also performed on two hundred datasets generated by bootstrap sampling with replacement. The median and the mean are 88.61\% and 87.77\% respectively.
Our results show that a small set of statistics of the
tight frame coefficients along certain orientations can serve as discriminative features for van Gogh paintings. It is more important to look at the tail distributions of such directional coefficients than mean values and standard deviations. It reflects a highly consistent style in van Gogh's brushstroke movements, where many forgeries demonstrate a more diverse spread in these features.}
\end{abstract}

\section{Introduction}
\label{introduction}

Art authentication is the identification of genuine paintings by famous
artists and detection of forgery paintings by imitators. The traditional way
in art authentication is to rely on the discerning eyes
and experience of experts who are dedicated in the work and life of
the artist(s). Physical means such as ultraviolet fluorescence \cite{liang2002apparatus},
infrared reflectography \cite{de1974note}, x-ray radiography \cite{newman1998applications},
painting sampling \cite{caddy2002forensic}, and canvas weave count \cite{berger1966weave}
have also been used for art authentication.
The term \emph{stylometry} refers to the application of statistical or
quantitative techniques for authorship and style evolution in literary arts \cite{pawlowski2004wincenty}. In the past decade, research
in stylometry for paintings has been benefited from the
rapid progress in image data acquisition technology.
By using high-resolution digital images of artists' collections,
image analysis researchers and art historians have engaged in cross-disciplinary
stylometric analysis of art paintings via computational techniques \cite{taylor1999fractal,lyu2004digital,li2004studying,berezhnoy2005authentic, berezhnoy2009automatic,hughes2012empirical,johnson2008image,li2012rhythmic,qi2013visual}.

Although many art authentication methods were proposed and
used, the authorship of many paintings is still questioned by experts,
with different art scholars having different opinions. Stylometry for
paintings, in particular, is still a long way from being a mature field,
even for paintings from well-known artists. Here, we propose a new stylometric
technique for art authentication of Vincent van Gogh paintings.
Our results on 79 paintings provided by van Gogh Museum and Kr\"oller-Muller
Museum show that our method is better than existing van Gogh paintings
authentication methods \cite{qi2013visual,johnson2008image,lyu2004digital}.

Stylometry is based on the assumption that there are some
distinctions in styles among different artists. Each artist
exhibited particular traces of natural style and habitual physical
movements when painting. Therefore, characteristics reflecting these
habits can be considered as features to identify the authorship of paintings.
In the past two decades, various specialized features have been used
in stylometric analysis, and many paintings are authenticated.
An early study was given by Taylor {\it et al.} in 1999 on fractal analysis
of Pollock's drip paintings \cite{taylor1999fractal}. They showed
that the fractal dimensions increased steadily through Pollock's
career and fractal analysis could be used as a quantitative and
objective technique for analyzing his paintings. In a 2004 paper by
Lyu {\it et al.} \cite{lyu2004digital}, the moment statistics of wavelet coefficients and the log error {in a} linear predictor are used as features to authenticate the drawings
by Pieter Bruegel the Elder.
In the same year, Li and Wang \cite{li2004studying}
put 2D multi-resolution Hidden Markov Model (HMM) in use to classify
paintings from some China's famous artists in different dynasty periods.
Later, Berezhnoy {\it et al.} \cite{berezhnoy2009automatic}
gave an orientation extraction technique based on circular
filters for brushstroke extraction. Recently, the moment statistics of 2-D
Empirical Mode Decomposition (EMD) coefficients
were used by Hughes {\it et al.} \cite{hughes2012empirical} for
stylometric analysis of drawings by Pieter Bruegel
the Elder and Rembrandt van Rijn. For each forgery in their dataset, a
binary classifier was trained based on this forgery together
with all but one genuine drawings. Then the left-out
genuine drawing was classified according to the trained classifier. However,
in the paper, there is no authentication done on the forgery drawings.

In 2008, three research groups from Penn State, Princeton, and Maastricht focusing on authenticating van Gogh paintings reported their analysis of van Gogh's
brushstrokes in \cite{johnson2008image}.
In the work of the Penn state group, the similarity among paintings
were assessed via texture and brushstroke geometry modeling.
The Princeton group applied the complex wavelet and Hidden Markov Tree (HMT)
for feature extraction, and then similarity distances between paintings were calculated
using the first few features ranked according to their effectiveness in distinguishing van Gogh's and non-van Gogh's patches. Finally a multidimensional scaling embeds the
paintings into a 3D space where the separation
of genuine paintings from forgery ones was done. Binary support vector machine
was used to determine the authorship by the Maastricht group. It
is based on the fact that the total energy, as calculated using the Gabor wavelet coefficients from the patches, was larger in the non-van
Gogh's paintings. These studies are quite encouraging as initial works
for identifying the authorship of van Gogh paintings.

In 2012, Li {\it et al.} \cite{li2012rhythmic} made an effort to extract those visually salient brushstrokes of van Gogh based on an integrative technique of both edge detection and clustering-based segmentation. With the extracted brushstrokes, some definitions of brushstroke features for art authentication were given in distinguishing van Gogh
paintings from forgeries. In their numerical test, they compared the brushstrokes obtained manually
with those extracted using their algorithm and showed that the combined brushstroke features  were consistent throughout van Gogh's works during his French periods (1886-1890).

More recently in 2013, Qi. {\it et al.} \cite{qi2013visual} use background selection and wavelet-HMT-based Fisher information distance for authorship and dating of impressionist and post-impressionist paintings. Two novel points were introduced in this work.
The first point is that background information is much more reliable than the details
of an intricate object which cannot represent the artist's natural style
because of multiple edits and corrections. Therefore they proposed to
seek out sections in the paintings that have been painted quickly
without too much modification. However, in their tests the labeling
of painting patches, either ``background'' or ``detail'', is done manually
by a non-expert. The second point is that an artist's style should be interpreted as a probability distribution over a set of possible textures,
and not just simply from the textures themselves.

For art authentication, the key point is to find the appropriate features
which give a good separation between the artist's paintings and those by his
imitators. In this paper we aim to find an appropriate measure so that the paintings
drawn by one artist, such as van Gogh, are much more similar than those by the imitators.
As in \cite{johnson2008image}, we first start by analyzing the brushstrokes in the
paintings by some analysis operators. Instead of using the variety of techniques such as
wavelets, EMD, HMM and HMT in \cite{lyu2004digital,hughes2012empirical,johnson2008image,qi2013visual},
here we propose to use
a special tight frame, called geometric tight frame \cite{li2011framelet}, to extract brushstroke information from the given paintings. 
Tight-frame transforms are redundant bases that can provide overcomplete 
but stable coding of directional variations \cite{cai2012image}.
The geometric tight frame we used has 18 filters that
give the first- and second-order differences in the horizontal, vertical and
diagonal directions in small neighborhoods. Therefore it can capture subtle oriented
variations in the texture of the paintings.

Next to find our features, we follow the moment statistics approach explored in \cite{lyu2004digital,hughes2012empirical} and propose to
use 3 simple statistics of the geometric tight frame coefficients as our features. 
 They are the mean of the coefficients, the standard deviation of the coefficients,
and the percentage of those coefficients that are
more than one standard deviation away from the mean.
That gives a total of 54 features for each painting. Then we
select the discriminatory features by a forward stage-wise
boosting procedure \cite{fawcett2004roc,hastie2009elements}.
It selects features by maximizing the area under the Receiver
Operating Characteristic (ROC) \cite{fawcett2004roc} curve such that van Gogh's paintings are highly concentrated
while the forgeries are widely spread.
We also used the leave-one-out cross-validation procedure to avoid overfitting while maximizing the amount of training data.

Once the features are selected, we use a simple thresholding rule to
authenticate the paintings.
Our test on the 79 paintings shows that we can achieve $86.08\%$ classification accuracy
and it reveals that
5 of the 54 features are predominant. By using just these 5 features,
we can get a $88.61\%$ classification accuracy which is highest so far
reported in the literature \cite{qi2013visual,johnson2008image,lyu2004digital}.
Evaluation of the five features is also performed on two
hundred 79-painting datasets generated by bootstrap sampling with replacement \cite{efron1994introduction}.
 The 95\% confidence interval is (78.48\%, 94.94\%) with median 88.61\%
and mean 87.77\%. Our results show that a small set of statistics of the tight frame coefficients along certain orientations in small neighborhood can serve as discriminative
features for van Gogh paintings. This reflects a highly consistent style in van Gogh's
brushstroke movements, where many forgeries demonstrate a more diverse spread
in these features.

Our proposed method, though tested only on van Gogh dataset, can easily
be applied to paintings by other artists. We hope that our method may
help art scholars to identify more digital evidences discriminating
different artists' paintings from forgeries.

This paper is organized as follows.  Section \ref{dataset}  describes the dataset we used for our art authentication. Section \ref{extraction} introduces how we
construct our features. Section \ref{fss} explains how we select
the most discriminatory features among the features we constructed. Section \ref{classification} describes how we used the selected features so obtained
to do the authentication. Section \ref{experiment} gives the numerical results. We draw a conclusion in Section \ref{conclusion}.

\section{Dataset}
\label{dataset}

Our dataset consists of 79 digitalized impressionist and post impressionist paintings provided to us by the Maastricht group \cite{johnson2008image}. They are
high-resolution color copies of paintings from the van Gogh museum and Kr\"{o}ller-Muller museum by professional scanners and are suitable for art research. These paintings vary in sizes, with the smallest one being 1452-by-833 pixels and the largest one 5614-by-7381 pixels. Among the 79 paintings, 64 paintings were drawn by van Gogh himself and the remaining 15 paintings were by his contemporaries. In the following,
we will abbreviate them as vG (van Gogh) and nvG (non-van Gogh) paintings respectively.
Sample images of vG and nvG
paintings are given in Figures \ref{fig:vg} and \ref{fig:nvg}.
It should be noted that the 15 forgeries are very similar to the 64 van Gogh's artworks,
with 6 of them historically attributed to van Gogh, but have been known to be forgeries now. Table \ref{nvg-confused} lists these six once-debatable paintings, which are regarded as difficult examples for stylometric analysis.
\begin{table}[h]
\begin{center}
\begin{tabular}{c|c|c}
 \small ID & \small Title &\small  Date and place \\ \hline
\small f233	& \small View of Montmartre with quarry	& \small Paris, Late 1886 \\
\small f253	& \small Still life with a bottle, Two glasses cheese and bread	& \small Paris, Spring 1886 \\
\small f253a &	\small A plate of rolls	& \small Paris, first half of 1887 \\
\small f278	& \small Vase with poppies, cornflowers peonies	& \small Paris, Summer 1886	 \\
\small f418	&\small  Family: On\'{e}sime Comeau/Marie-Louise Duval 	& \small Jan., 1890	 \\
\small f687	& \small Reaper with sickle (after Millet) 	 & \small Saint-Remy Sep., 1889
\end{tabular}
\vspace*{10pt}
\caption{The 6 paintings which were once wrongly attributed to van Gogh in history.} \label{nvg-confused}
\end{center}
\end{table}
We make a note about the boundary of the paintings here. As pointed out in Qi {\it et al.} \cite{qi2013visual}, the edges of the canvas in the paintings may not
be useful information for art authentication,
and hence we have excluded these edges in our numerical experiments. More precisely,
for each painting in the dataset, we crop off 100 pixels from its four sides, and
use only the interior of the image in our numerical tests.

In the following we focus on  finding a small set of features to automatically classify the 79 paintings into vG and nvG. With a small set of samples, to avoid the problem of overfitting, we adopt the leave-one-out test method \cite{geisser1993predictive}. Then
a forward stage-wise boosting procedure \cite{fawcett2004roc,hastie2009elements}
is used to construct a small set of features such that in such feature
space those vG's are highly concentrated in a cluster while
the nvG's are mostly spread away from such a cluster.
More precisely, we consider the art authentication problem as an outlier
detection problem, where vG's are the normal data and nvG's are the outliers.
In the next three sections we introduce the methodology to achieve this goal
and here is the outline of these three sections.

Section \ref{extraction} introduces our approach of extracting features from the
paintings. It is based on an 18-filter geometric tight frame and 3 simple statistics
so that 
each painting is represented by a 54-dimensional vector after feature extraction. The accuracy of our method is tested by a leave-one-out test
scheme which uses one painting from the original dataset as the test painting
and the remaining 78 paintings as the training data.
  In Section \ref{fss},
we describe how the forward stage-wise boosting procedure is used to select 5
important features among the 54 features. In Section \ref{classification}, we focus on the classification
of the test painting. In Subsection \ref{rule}, we describe how to construct the classification rule once the 5 features are selected and in Subsection \ref{test}
how to use the rule to classify the test painting. This procedure is repeated 79 times
such that each painting in the dataset is tested once. The classification accuracy
of our method is measured by the results of these 79 tests.

\section{Feature Extraction}
\label{extraction}
Tight frames have been used successfully in different
applications in image processing \cite{dong2010mra, chan2004tight, cai2012image}.
The geometric tight frame we use to analyze the brushstrokes in our paintings
is proposed in \cite{li2010multiframe, li2011framelet} and it
can capture the first- and second-order differences in the horizontal, vertical
and diagonal directions in every small neighborhood of the paintings.
As discovered in \cite{lyu2004digital,hughes2012empirical,qi2013visual}, statistical properties of quantities such as wavelet coefficients, EMD coefficients or HMT-parameters
are useful in authenticating paintings by various artists. Here we combine the two ideas
and propose to use some simple statistics of the geometric tight frame coefficients of
each painting as our features. We will see from the numerical results
in Section \ref{experiment} that our features can
do a good job in capturing the rapid, rhythmic, and vigorous brushstroke movements of van Gogh, and hence discriminating his paintings from those forgeries of his contemporaries.

In the following, we introduce the geometric tight frame and
the statistics used in this paper.

\subsection{Geometric tight frame}
\label{decomposition}
The geometric tight frame we use has 18 filters $\tau_0,\tau_1,\cdots, \tau_{17}$:
\begin{eqnarray}
\tau_0=\frac{1}{16}\begin{bmatrix}
1      & 2  & 1      \\
2      & 4  & 2 \\
1      & 2  & 1
\end{bmatrix},&
\tau_1=\frac{1}{16}\begin{bmatrix}
1      & 0  & -1      \\
2      & 0  & -2 \\
1      & 0  & -1
\end{bmatrix},&
\tau_2=\frac{1}{16}\begin{bmatrix}
1      & 2  & 1      \\
0      & 0  & 0 \\
-1      & -2  & -1
\end{bmatrix}, \nonumber \\
\tau_3=\frac{\sqrt{2}}{16}\begin{bmatrix}
1      & 1  & 0      \\
1      & 0  & -1 \\
0      & -1 & -1
\end{bmatrix},&
\tau_4=\frac{\sqrt{2}}{16}\begin{bmatrix}
0      & 1  & 1      \\
-1     & 0  & 1 \\
-1     & -1 & 0
\end{bmatrix},&
\tau_5=\frac{\sqrt{7}}{24}\begin{bmatrix}
1      & 0  & -1      \\
0      & 0  & 0 \\
-1     & 0  & 1
\end{bmatrix}, \nonumber\\
\tau_6=\frac{1}{48}\begin{bmatrix}
-1      & 2  & -1      \\
-2      & 4  & -2 \\
-1      & 2  & -1
\end{bmatrix},&
\tau_7=\frac{1}{48}\begin{bmatrix}
-1     & -2  & -1      \\
2      & 4   & 2 \\
-1     & -2  & -1
\end{bmatrix},&
\tau_8=\frac{1}{12}\begin{bmatrix}
0      & 0   & -1      \\
0      & 2   & 0 \\
-1     & 0   & 0
\end{bmatrix}, \nonumber\\
\tau_9=\frac{1}{12}\begin{bmatrix}
-1     & 0   & 0      \\
0      & 2   & 0 \\
0      & 0   & -1
\end{bmatrix},&
\tau_{10}=\frac{\sqrt{2}}{12}\begin{bmatrix}
0      & 1   & 0      \\
-1     & 0   & -1 \\
0      & 1   & 0
\end{bmatrix},&
\tau_{11}=\frac{\sqrt{2}}{16}\begin{bmatrix}
-1     & 0  & 1      \\
2      & 0  & -2 \\
-1     & 0  & 1
\end{bmatrix},\nonumber\\
\tau_{12}=\frac{\sqrt{2}}{16}\begin{bmatrix}
-1     & 2  & -1      \\
0      & 0  & 0 \\
1      & -2 & 1
\end{bmatrix},\ &
\tau_{13}=\frac{1}{48}\begin{bmatrix}
1      & -2  & 1      \\
-2     & 4   & -2 \\
1      & -2  & 1
\end{bmatrix},\ &
\tau_{14}=\frac{\sqrt{2}}{12}\begin{bmatrix}
0      & 0   & 0      \\
-1     & 2   & -1  \\
0      & 0  & 0
\end{bmatrix}, \nonumber\\
\tau_{15}=\frac{\sqrt{2}}{24}\begin{bmatrix}
-1     & 2   & -1      \\
0      & 0   & 0 \\
-1     & 2  & -1
\end{bmatrix},\ &
\tau_{16}=\frac{\sqrt{2}}{12}\begin{bmatrix}
0      & -1  & 0      \\
0      & 2   & 0 \\
0      & -1  & 0
\end{bmatrix},\ &
\tau_{17}=\frac{\sqrt{2}}{24}\begin{bmatrix}
-1     & 0   & -1      \\
2      & 0   & 2 \\
-1     & 0   & -1
\end{bmatrix}, \nonumber \\ \label{filters}
\end{eqnarray}
see \cite{li2010multiframe, li2011framelet}.
We note that $\tau_0$ is the low-pass filter. The filters
$\tau_1$, $\tau_2$, $\tau_3$ and $\tau_4$ are the Sobel operators in the vertical, horizontal, $-\frac{\pi}{4}$, and $\frac{\pi}{4}$ directions, respectively whereas the filters
$\tau_8$, $\tau_9$, $\tau_{14}$, $\tau_{15}$, $\tau_{16}$ and $\tau_{17}$ are the
second-order difference operators in different directions.

 Given the $i$-th color painting, $1\le i \le 79$, with $m_i$-by-$n_i$ pixels,
we represent its grey-scale intensity by an $m_i$-by-$n_i$ matrix $P_i$. Then
we convolve $P_i$ with each $\tau_j$, $0\le j\le 17$, to get the corresponding
$m_i$-by-$n_i$ tight frame coefficient matrices:
\begin{eqnarray}
&& A^{(i,j)}=P_i \ast \tau_j
 =\begin{pmatrix}
     a^{(i,j)}_{1,1} & \cdots & a^{(i,j)}_{1,n_i}   \\
      \vdots& & \vdots\\
      a^{(i,j)}_{m_i,1} & \cdots & a^{(i,j)}_{m_i,n_i}
\end{pmatrix}, \qquad
  1\le i \le 79, \ 0 \le j \le 17. \label{coefficients}
\end{eqnarray}
Therefore, there are 18 corresponding coefficient matrices for each painting after the decomposition by the geometric tight frame. We remark that we only use one level of
the tight frame transform without any down-sampling.
Our numerical result shows that using 2 levels of tight frame transform gives a bad classification accuracy, see Table \ref{result}. Moreover,
it increases the number of features significantly.

\subsection{Three statistics} \label{def:6stat}
Moment statistics has been used successfully to extract features in art authentication, see \cite{lyu2004digital,hughes2012empirical,mao2010new}. In \cite{hughes2012empirical}, the moment statistics of the ``outlier pixels'', defined as those that are {\it greater than} the mean plus one standard deviation,
are also considered as features. Thus here we propose to use the
following three statistics as features. They are (i) the mean
of the entries in the coefficient matrix, (ii)
 the standard deviation of the entries in the coefficient matrix,
 and (iii) the percentage of the ``tail entries" which are those
 entries that are {\it more than} one standard deviation from the mean.
To be precise, given a coefficient matrix $A^{(i,j)}$ in (\ref{coefficients}) with entries
$a^{(i,j)}_{l,k}$,
the three statistics are defined as follows:
\begin{itemize}
\item [(\romannumeral1)] the mean of $A^{(i,j)}$: $$\mu^{(i,j)}=\frac{1}{m_in_i}\sum^{m_i}_{l=1}\sum^{n_i}_{k=1}a^{(i,j)}_{l,k},$$
\item[(\romannumeral2)] the standard deviation of $A^{(i,j)}$: $$\sigma^{(i,j)}=\left(\frac{1}{m_in_i-1}\sum\limits^{m_i}_{l=1}
\sum^{n_i}_{k=1}\left(a^{(i,j)}_{l,k}
-\mu^{(i,j)}\right)^2\right)^\frac{1}{2},$$
\item[(\romannumeral3)] the percentage of the tail entries
${p}^{(i,j)}={\#(\hat{A}^{(i,j)})}/{(m_in_i)}$.
\end{itemize}

\vspace{2mm}\noindent
Here $\#(\hat{A}^{(i,j)})$ is the number of nonzero entries in  the tail matrix $\hat{A}^{(i,j)}$ which is defined by
$$
\hat{a}^{(i,j)}_{l,k}=\left\{ \begin{array}{cl}
a^{(i,j)}_{l,k}, & \mathrm{if}\ |a^{(i,j)}_{l,k}-\mu^{(i,j)}|>\sigma^{(i,j)},\\
0, & \mathrm{otherwise}.
\end{array} \right.
$$
Thus the feature vector of the $i$-th painting is represented by
\begin{equation}
 \left[\mu^{(i,0)},\cdots, \mu^{(i,17)},{\sigma}^{(i,0)},\cdots,{\sigma}^{(i,17)}, {p}^{(i,0)},
\cdots,{p}^{(i,17)} \right] \in {\mathbb R}^{54}. \label{fv}
\end{equation}

In summary, we have 79 paintings and 54 features.
The accuracy of our method is tested by a leave-one-out procedure
where a painting, say $P$, in the dataset is used as the testing data
and the remaining 78 paintings in the dataset are used as the training data
to  select a small feature subset ${\cS}\subseteq\{1,2, \ldots, 54\}$. Then ${\cS}$
is used to train a classifier to test the left-out painting $P$.
The next section describes how to select $\cS$.

\section{Forward stage-wise feature selection procedure} \label{fss}
Considering the highly rhythmic brushstroke movements of van Gogh, it is 
unlikely that all the 54 features are discriminative between van Gogh
and his contemporaries. If we include some noisy features in our
classification task, the accuracy will deteriorate. Therefore, in this section we develop a feature selection
method based on a forward stage-wise rank boosting \cite{hastie2009elements} to boost the discriminating power of our feature sets.
Such a feature selection plays an indispensable role in our method,
which not only greatly improves our classification accuracy but also
leads to interpretable models---where with only five features we can
reach a classification accuracy of 88.61\%. 
In this section, we describe our procedure to
select a good feature subset from the given training dataset.

The way we discriminate van Gogh's paintings from the forgeries is based on
 the assumption that van Gogh exhibits highly consistent brushstroke movements in some of the texture features. Therefore under these features van Gogh's paintings will be
highly  concentrated toward some center points while forgeries are spread as outliers.
To be more precise, let the training dataset
be ${\cal X}=\{\mathbf{x}_1,\cdots,\mathbf{x}_{78}\}$ and
$$
X=
\left[
\begin{array}{c}
\mathbf{x}_1 \\
\vdots \\
\mathbf{x}_{78}
\end{array}
\right] \in \mathbb{R}^{78 \times 54}
$$
be the data matrix of  ${\cal X}$ and $\tilde{X}$ be the normalization of $X$
such that each column in $\tilde{X}$ has a unit standard deviation. Let
$\{1,\ldots,78\}=\T_{\mathrm{vG}}\cup \T_{\mathrm{nvG}}$  where $\T_{\mathrm{vG}}$ (respectively $\T_{\mathrm{nvG}}$) denotes the set of vG (respectively nvG)
paintings in ${\cal X}$.
For any feature subset ${\cal F}$, denote $|{\cal F}|$ the number of elements in the set ${\cal F}$ and ${\cal F}=\{i_1,\cdots,i_{|{\cal F}|}\}$. Define
$\tilde{X}_{j{\cal F}}=(\tilde{X}_{ji_1},\cdots,\tilde{X}_{ji_{|{\cal F}|}})$,
i.e. $\tilde{X}_{j{\cal F}}$ is the $j$-th row of $\tilde{X}$ restricted onto
the index set $\cal F$.
Then we define the \emph{vG center} w.r.t. ${\cal F}$ as the mean vector of vG on
${\cal F}$, i.e.
\begin{equation}
\label{centervG}
{\mathbf{c}}^{\cal F}=\frac{1}{|\T_{\mathrm{vG}}|}
\sum_{j\in \T_{\mathrm{vG}}}
\tilde{X}_{j{\cal F}}.
\end{equation}
With this we define the distance between the $j$-th painting in ${\cal X}$
and the vG center ${\mathbf{c}}^{\cal F}$ by
\begin{equation}
\label{distance}
d^{{\cal F}}_j=\|\tilde{X}_{j\F}-\mathbf{c}^{\F}\|_2, \quad 1 \le j \le 78.
 \end{equation}
For $\F$ to be a good feature set, $d^{{\cal F}}_j$ should be small
for $j\in \T_{\mathrm{vG}}$ and large for $j\in \T_{\mathrm{nvG}}$, i.e. nvG
should be far from the vG center and regarded as {outliers}.

To quantitatively measure any given $\F$, we use the theory of the ROC curve which has been widely used in literature \cite{zweig1993receiver,briggs2008skill,huang2007evaluating,
moskowitz2004quantifying,davis2006relationship,fawcett2004roc}.
Let us sort $\{d^{\F}_j\}_{j=1}^{78}$ in (\ref{distance}) in  an ascending order such that
$d^{\F}_{j_1} \leq d^{\F}_{j_2} \leq \ldots \leq d^{\F}_{j_{78}}$.
For any number $\rho$ (smaller than $d^{\F}_{j_1}$, larger than $d^{\F}_{j_{78}}$ or in between $d^{\F}_{j_1}$ and $d^{\F}_{j_{78}}$),
we can use it as a binary classifier to label all the paintings in ${\cal X}$.
From that we can determine the true positive rate
and the false positive rate w.r.t. $\rho$ (see (\ref{tpr}) for the definitions
of the rates). By plotting the true positive
rate versus the false positive rate for different $\rho$,
we obtain the ROC curve w.r.t. $\F$. Then we can compute the
area under the ROC curve $\operatorname{AUC}(\F)$. Notice that
the larger $\operatorname{AUC}(\F)$ is, the better $\F$ is as
more vG are close to the vG center and more nvG
are far from the vG center \cite{fawcett2004roc}. In the maximal case that $\operatorname{AUC}(\F)=1$, the nvG's distances are all greater than
any vG's distances and
there is a suitable $\rho$ that can classify all paintings correctly.

Therefore, the best feature subset $\F$ would be the one that maximizes $\operatorname{AUC}(\F)$.
However this is intractable due to the curse of dimensionality with an exponential
blow-up of computational complexity. Thus we adopt the following forward stage-wise approach
(see \cite{hastie2009elements}) to maximize $\operatorname{AUC}(\F)$. We start from the empty set $\F^{(0)}=\emptyset$
and iterate. Suppose at the $j$-th iteration, we already find $\F^{(j)}$. Then
in the $(j+1)$-th iteration, we greedily select the next feature by
\begin{eqnarray*}
l_{j+1}=\arg\max_{l\not \in \F^{(j)}}\operatorname{AUC}\left(\F^{(j)}\cup\{l\}\right),
\end{eqnarray*}
and update $\F^{(j+1)}=\F^{(j)}\cup\{l_{j+1}\}$.
In our numerical experiment, we stop at the fifth iteration. Thus the resulting
feature set for ${\cal X}$ is ${\cS}=\F^{(5)}$, and it has 5 features.

Note that for a dataset of $n$ paintings with $f$ features
(ours has $f=54$ and $n=78$) such a forward procedure has a computational cost of $O(nj(f-j))$ at the $j$-th iteration.

\section{Classification and Validation}
\label{classification}
In the following, we give the classification rule and how to use the classification rule to determine whether the left-out painting is genuine or fake.

\subsection{Classification rule}
\label{rule}
Given the selected features in ${\cS}$,
we already have  (see (\ref{centervG}) and (\ref{distance})) the vG center
$\mathbf{{c}}^{\cal G}$ corresponding to ${\cS}$ and
the distance $d_i^{\cal G}$ of the $i$-th painting
in ${\cal X}$ to $\mathbf{{c}}^{\cal G}$, for $1\le i \le 78$.
If $\cS$ is a good feature set, we expect $d_i^{\cal G}$ to be small for vG and large for nvG. Therefore our classifier is based on a simple threshold $\delta$, such that paintings
with ${d}_i^{\cal G}< \delta$ will be classified into vGs, or into nvGs if otherwise.

To determine $\delta$, a natural choice will be to maximize the classification accuracy (see (\ref{accuracy}) for the definition).
To be precise, let us sort $\{d_i^{\cal G}\}_{i=1}^{78}$ as ${d}_{i_1}^{\cal G}\le {d}_{i_2}^{\cal G}\le\cdots\le {d}_{i_{78}}^{\cal G}$ and define $(e_1,\cdots,e_{78})=({d}_{i_1}^{\cal G},\cdots,{d}_{i_{78}}^{\cal G})$.
Let  $(b_1,\cdots,b_{78})$ be the labels of the paintings in ${\cal X}$,
i.e.  $b_j=1$ if the $i_j$-th painting is vG and $-1$ otherwise.
For any threshold inside the interval $({d}_{i_{j-1}}^{\cal G}, {d}_{i_{j}}^{\cal G})$,
its accuracy is
$$
\epsilon_{j}=\frac{\sum\limits^{j-1}_{l=1}\left|\{l:b_l=1\}\right|
+\sum\limits^{78}_{l=j}\left|\{l:b_l=-1\}\right|}{78}, \quad j=1,\cdots, 79.
$$
We should therefore choose $j$ to be the one that maximizes $\epsilon_j$.
But as such $j$ may not be unique, we choose
$j^*=\max\{\arg\max_j\ \epsilon_{j}\}$,
and then the classification threshold is defined to be $\delta=\frac{e_{j^*-1}+e_{j^*}}{2}$.

\subsection{Classifying the left-out painting}
\label{test}
With the classification threshold $\delta$ defined, now we are ready
to classify the left-out painting $P$.
Let $\mathbf{z}\in \mathbb{R}^{5}$ be the feature vector extracted from $P$ according to the
feature set $\cS$. Then we normalize $\mathbf{z}$ to get $\tilde{\mathbf{z}}$,
i.e. we divide each entry in $\mathbf{z}$ by the corresponding
column standard deviation of $X$. Then the distance between the test painting $P$ and
the vG center $\mathbf{c}^{\cal G}$ is
$d=\|\tilde{\mathbf{z}}-{\mathbf{c}}^{\cal G}\|_2$.
We now classify $P$ as vG if $d<\delta$, or as nvG if otherwise.

Since we have 79 paintings in the dataset, the leave-one-out cross-validation procedure
described in Sections \ref{extraction}--\ref{classification}
is repeated 79 times such that each painting in the dataset is
tested as a left-out painting once. In particular,
our method is used 79 times to authenticate each of the 79 paintings
in our dataset once. The classification accuracy of the method is defined to be
the percentage of correct classifications (either genuine or forgery) in these
79 tests. We will carry out these tests in Section \ref{experiment}.

\section{Experimental Results}
\label{experiment}
This section gives the experimental results.
In Subsection \ref{nr}, we give the result for our method and compare it with those of others. Subsection \ref{analysis} identifies the
most discriminatory features obtained from our methods.
In Subsection \ref{evaluation}, we statistically evaluate our method and also
the most discriminatory features selected by our method.

\subsection{Results comparison}
\label{nr}
True positive (TP) is defined as the number of correct detection in
the vG test cases and true negative (TN) is the number of correct detection in the
nvG test cases. Then the true positive rate (TPR), true negative rate (TNR),
and the classification accuracy are defined as
\begin{eqnarray}
\mathrm{TPR}=\frac{\mathrm{TP}}{\mathrm{vG\ number}}, &\quad\ & \mathrm{TNR}=\frac{\mathrm{TN}}{\mathrm{nvG\ number}}, \label{tpr} \\
{\rm \ classification \ accuracy} & = & \frac{\mathrm{TP}+\mathrm{TN}}{\mathrm{total\ number}}.
\label{accuracy}
\end{eqnarray}

We have performed our method
on the 79 paintings in our dataset. Recall that there are 64 vG and 15 nvG paintings.
In our experiment, 60 out of the 64 van Gogh paintings are detected correctly
by our method as genuine
(i.e. $\mathrm{TP}=60$) and 8 out of the 15 imitations are detected correctly as forgery
(i.e. $\mathrm{TN}=8$). Therefore $(60+8)/79=86.08\%$ of these paintings
are classified correctly by our method.
{Table} \ref{result} gives the classification results by our method as
compared with some previous methods. We emphasize that there are 3 datasets: our dataset (64 vG, 15 nvG), IP4AI1 (54 vG, 11 nvG) and IP4AI2 (65vG, 15 nvG). The numbers reported on IP4AI1 and IP4AI2 are all computed by Qi {\it et al.} \cite{qi2013visual}.

We see from {Table} \ref{result} that our method gives the highest true positive
rate and a rather good classification rate.
In the second row of {Table} \ref{result}, we list the result if no feature selection
is done and all 54 features are used in our method. We see that the result is bad and
it is indeed necessary to perform feature selection to exclude those features which
are irrelevant or noisy and hence may obscure accurate classification. In the third row of Table \ref{result}, we give the results
when the 2-level geometric tight frame is used instead of 1-level.
We see that the classification accuracy is also bad, indicating
features with bigger neighborhood are not good in discriminating van Gogh's paintings.
Note that, in the 2-level case, we have 105 features to begin with instead of 54.

\begin{table}
\footnotesize
\begin{center}
\begin{tabular}{ c|c|c|c|c }
dataset &methods &TPR (TP)& TNR (TN) & accuracy (TP$+$TN) \\ \hline
\multirow{2}{*}{our dataset }& our method & {\bf 93.75\%} (60)& 53.33\% (8)&  86.08\% (68)\\\cline{2-5}
& All 54 features used & 89.06\% (57)& 0.00\% (0) & 72.15\% (57) \\\cline{2-5}
(64 vG, 15 nvG)&2-level tight frame & 84.36\% (54)& 26.67\% (4) & 73.42\% (58)\\ \hline\hline
\multirow{2}{*}{IP4AI1}  &WHMT\_FI& 92.59\% (50) & 63.64\% (7) & {\bf 87.69\%} (57)\\  \cline{2-5}
 &HMT with MDS& 90.74\% (49) & 36.36\% (4) & 81.54\% (53)\\  \cline{2-5}
(54 vG, 11 nvG)& LP with MDS&96.30\% (52) &36.36\% (4) & 86.15\% (56)\\ \hline\hline
\multirow{2}{*}{IP4AI2} &WHMT\_FI& 87.69\% (57)& {\bf 73.33\%} (11)& 85.00\% (68) \\  \cline{2-5}
 &HMT with MDS& 86.15\% (56)& 26.67\% (4) & 75.00\% (60) \\   \cline{2-5}
 (65 vG, 15 nvG)&LP with MDS&84.62\% (55) &40.00\% (6) & 76.25\% (61)
\end{tabular}
 \caption{Comparison of the classification results.}\label{result}
\vspace{10pt}
\end{center}
\end{table}

\begin{figure}[h]
\begin{center}
\subfigure[f249]{
\label{Fig.sub11}
\includegraphics[width=0.20\textwidth]{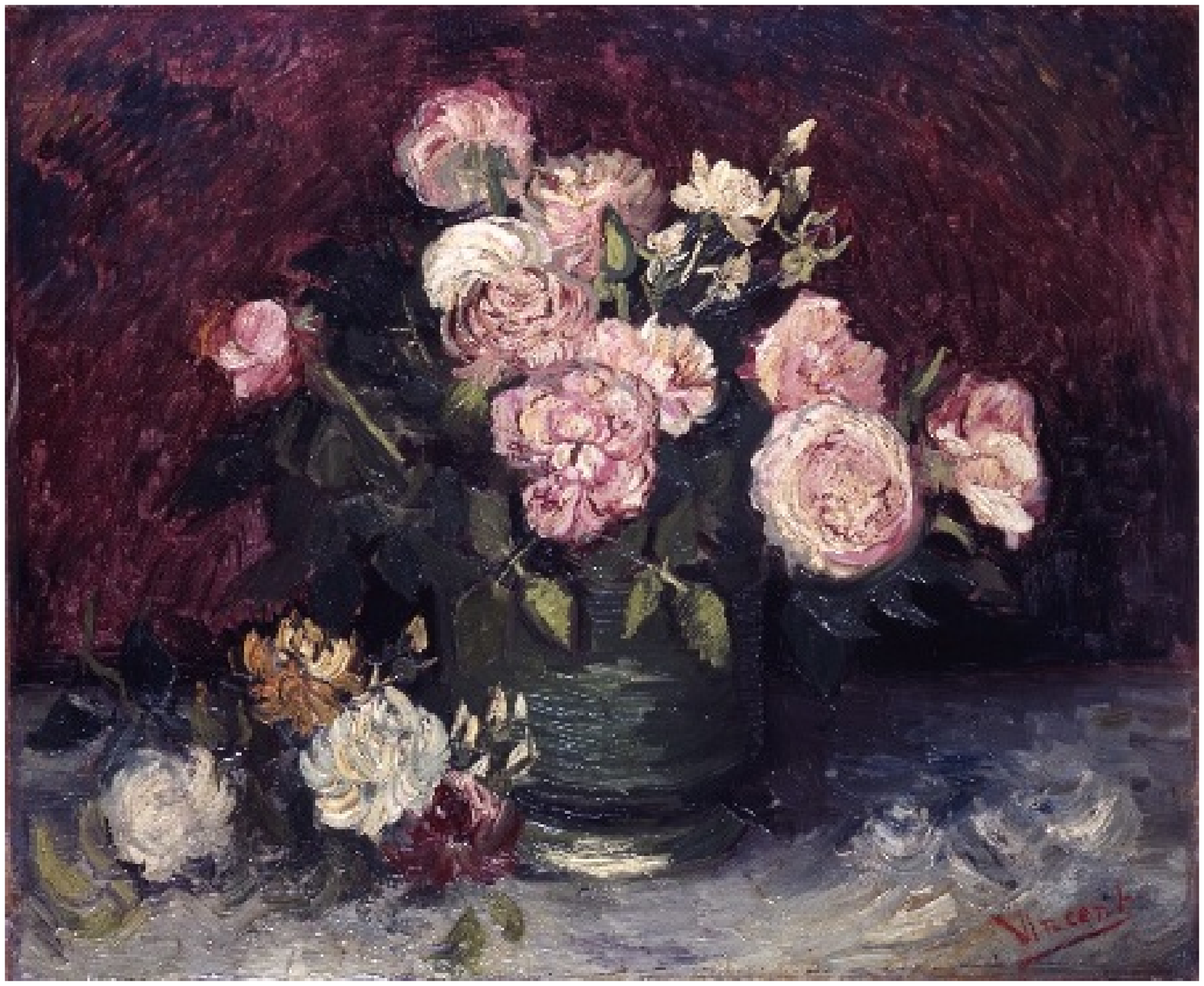}}
\subfigure[f371]{
\label{Fig.sub12}
\includegraphics[width=0.20\textwidth]{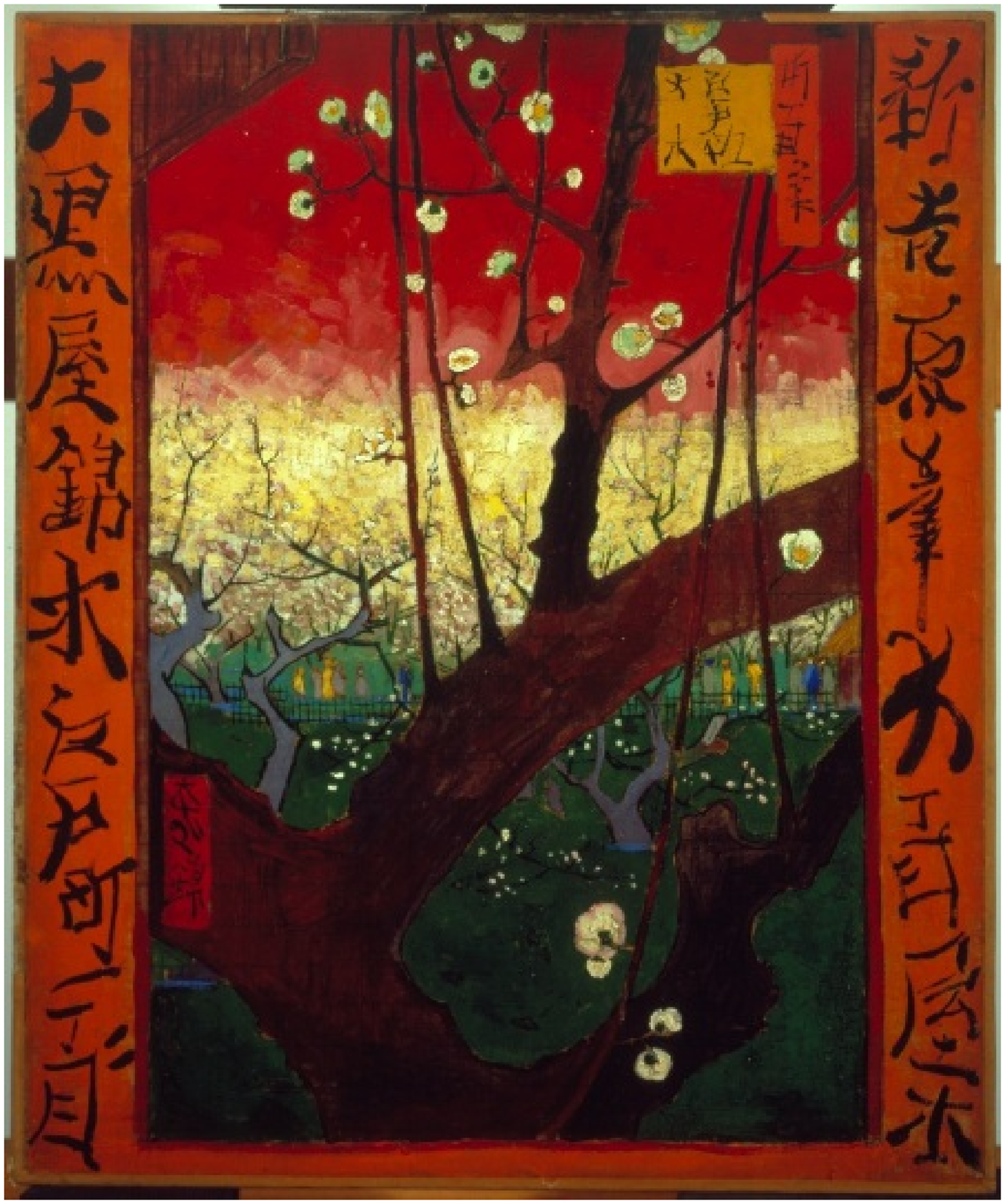}}
\subfigure[f522]{
\label{Fig.sub13}
\includegraphics[width=0.20\textwidth]{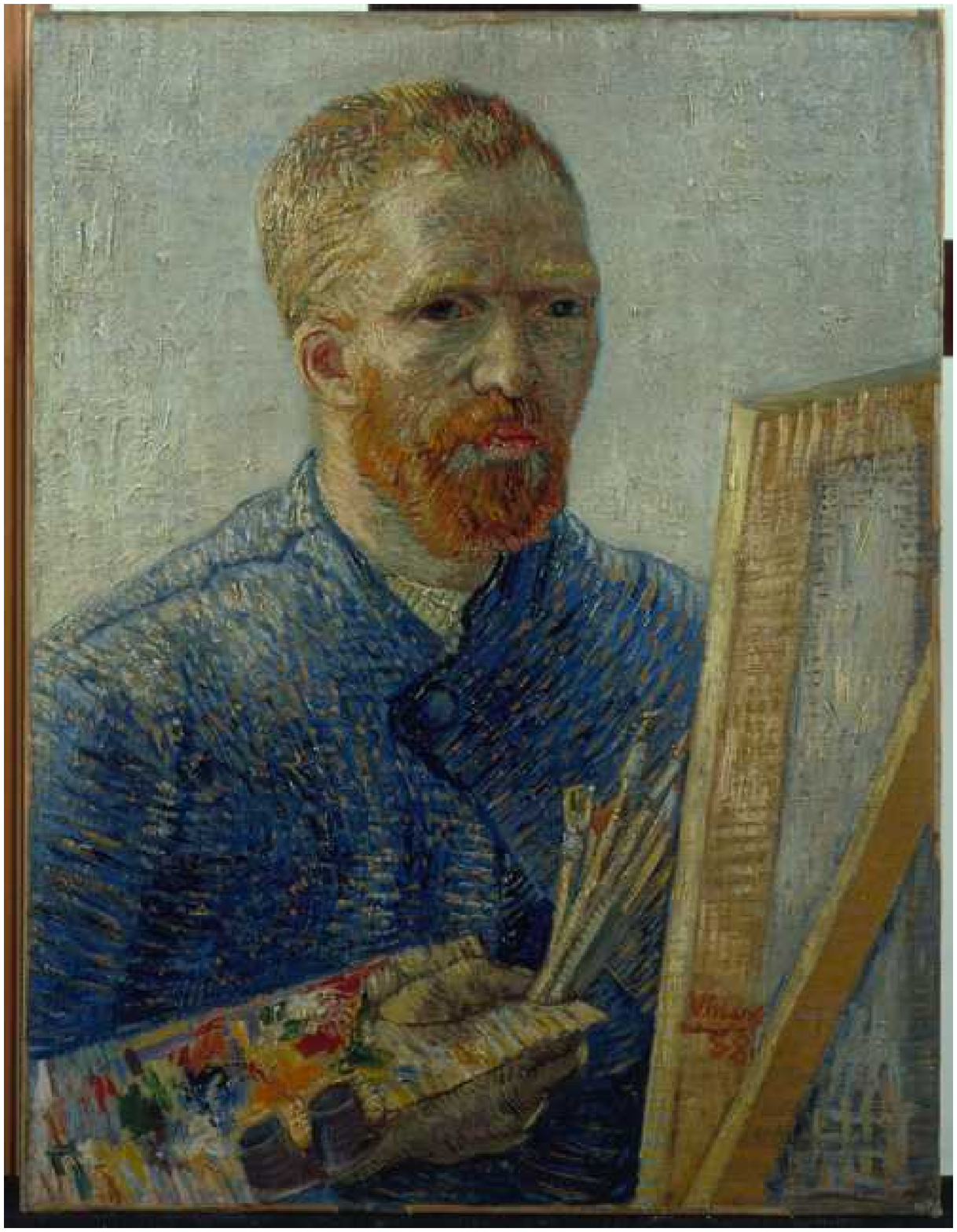}}
\subfigure[f752]{
\label{Fig.sub14}
\includegraphics[width=0.20\textwidth]{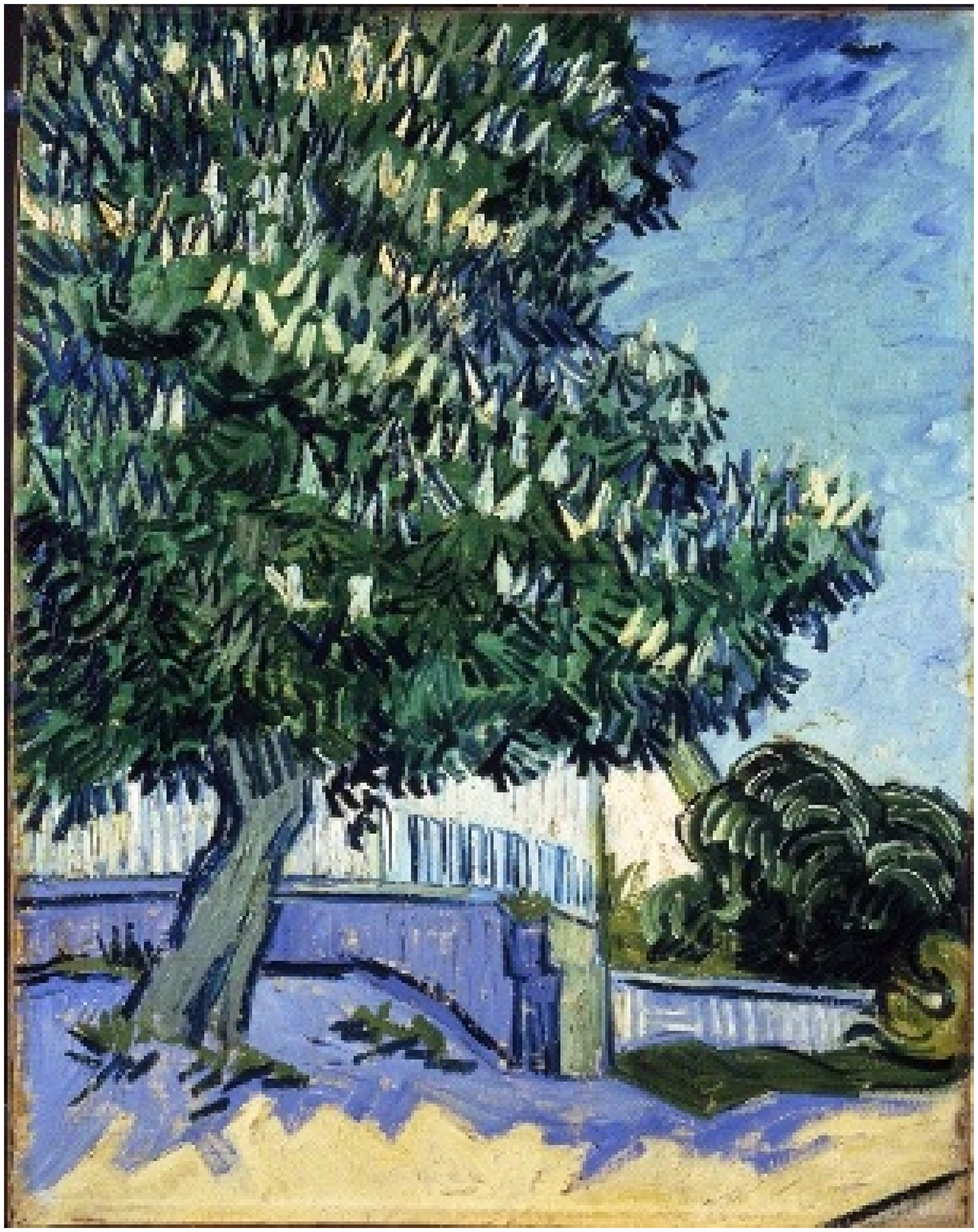}}
\caption{The four van Gogh's paintings detected wrongly by our method.}
\label{fig:vg}
\end{center}
\end{figure}

\begin{figure}[h]
\begin{center}
\subfigure[f253a]{
\label{Fig.sub1}
\includegraphics[width=0.20\textwidth]{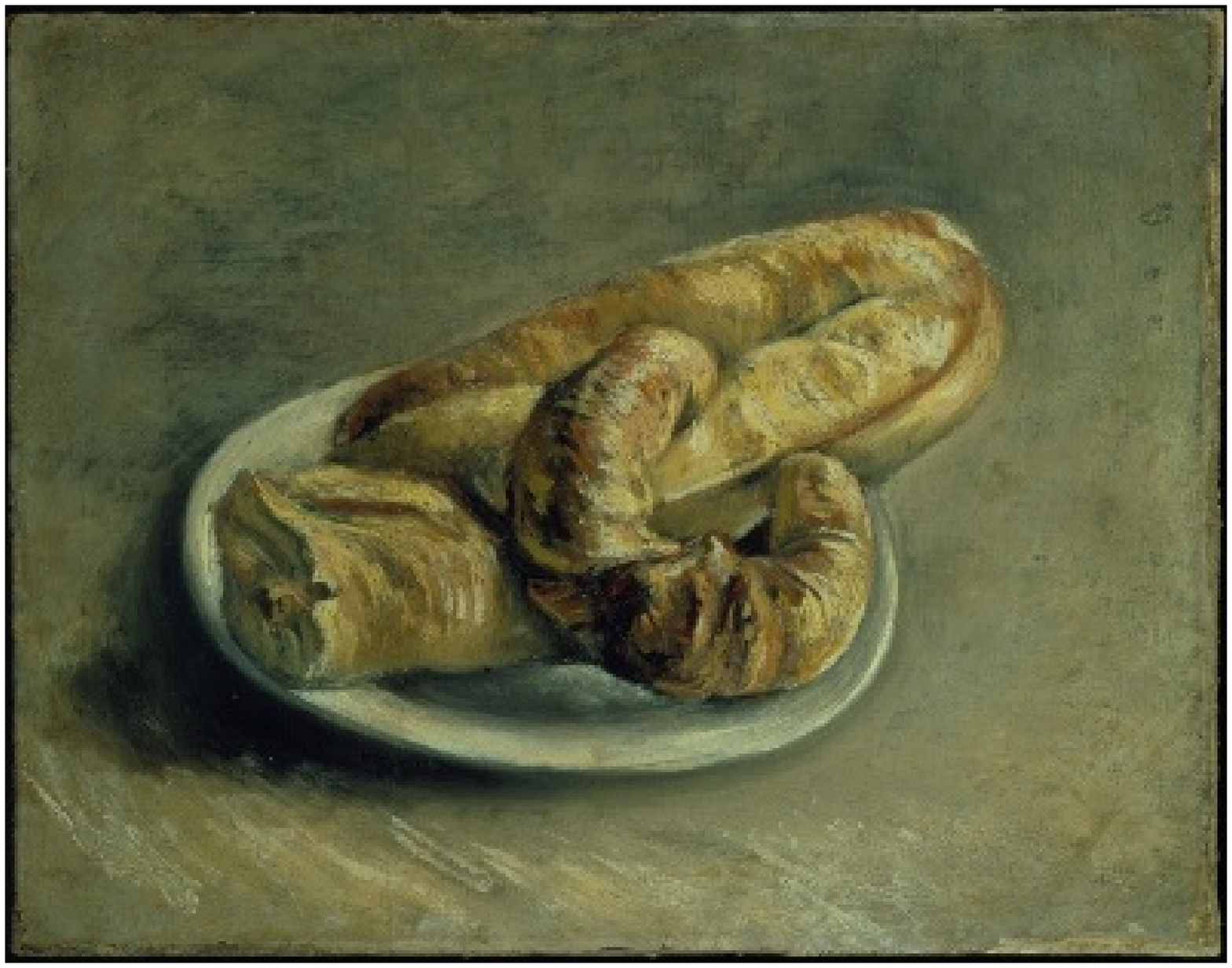}}
\subfigure[f418]{
\label{Fig.sub2}
\includegraphics[width=0.20\textwidth]{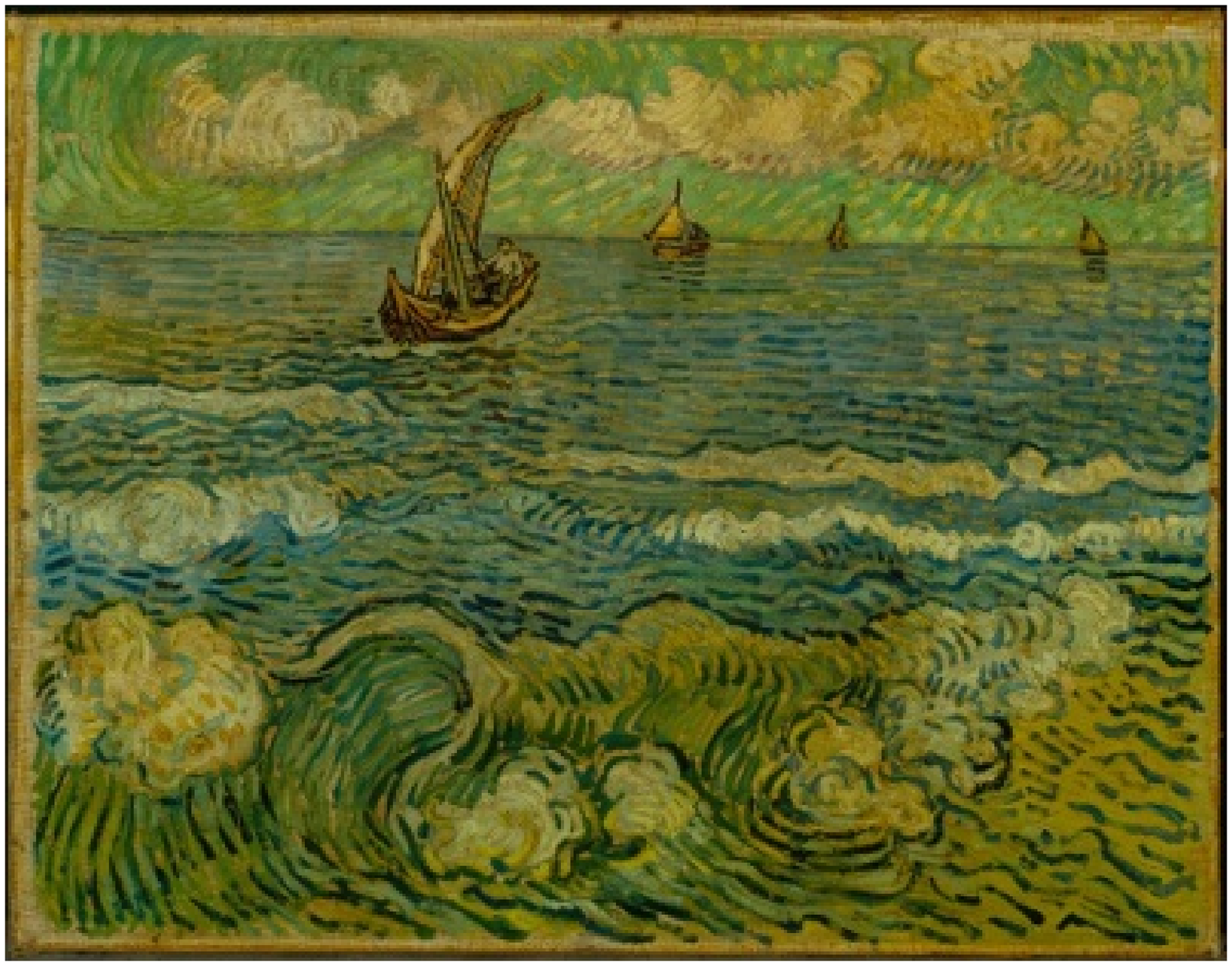}}
\subfigure[f687]{
\label{Fig.sub3}
\includegraphics[width=0.18\textwidth]{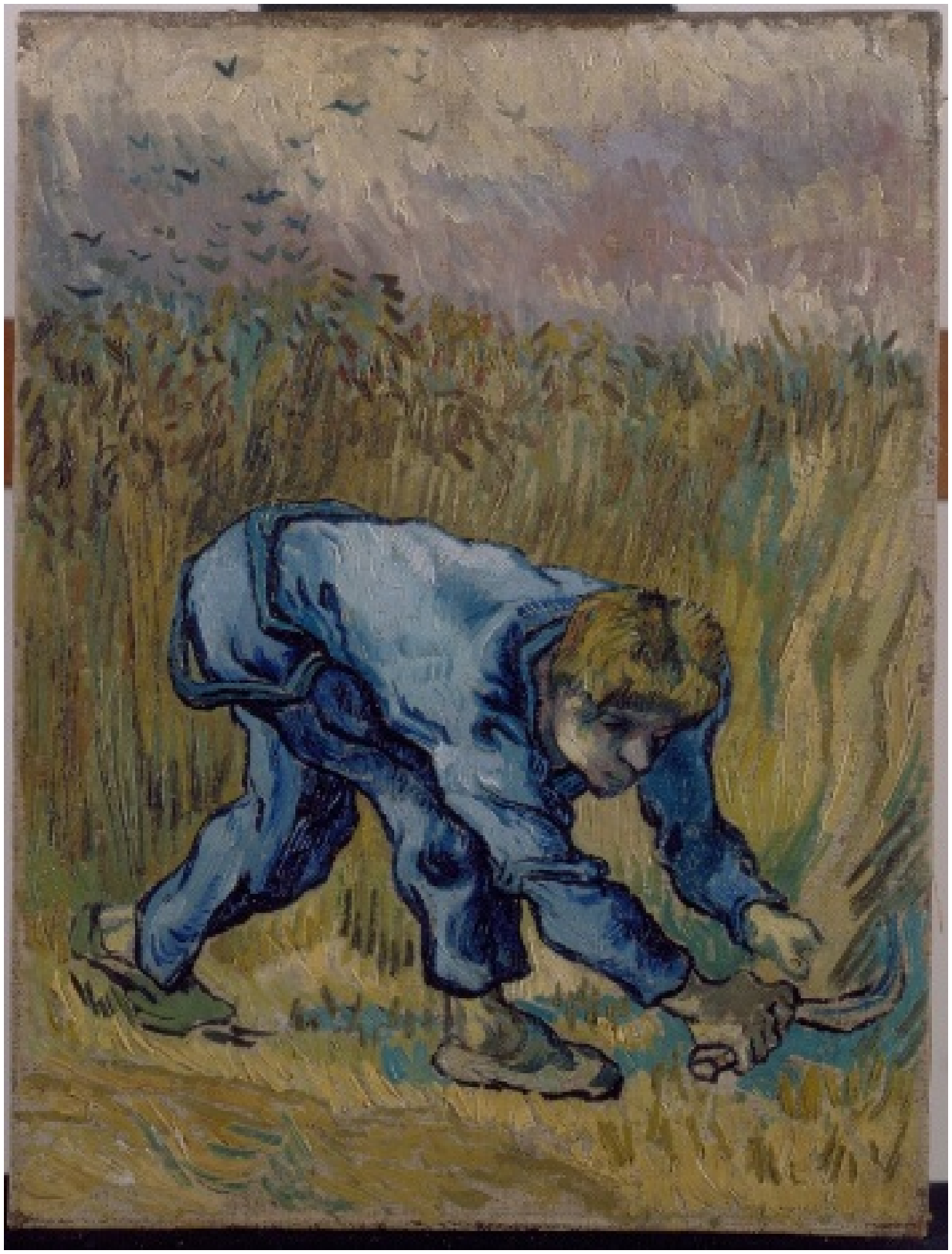}}\\
\subfigure[s503]{
\label{Fig.sub4}
\includegraphics[width=0.20\textwidth]{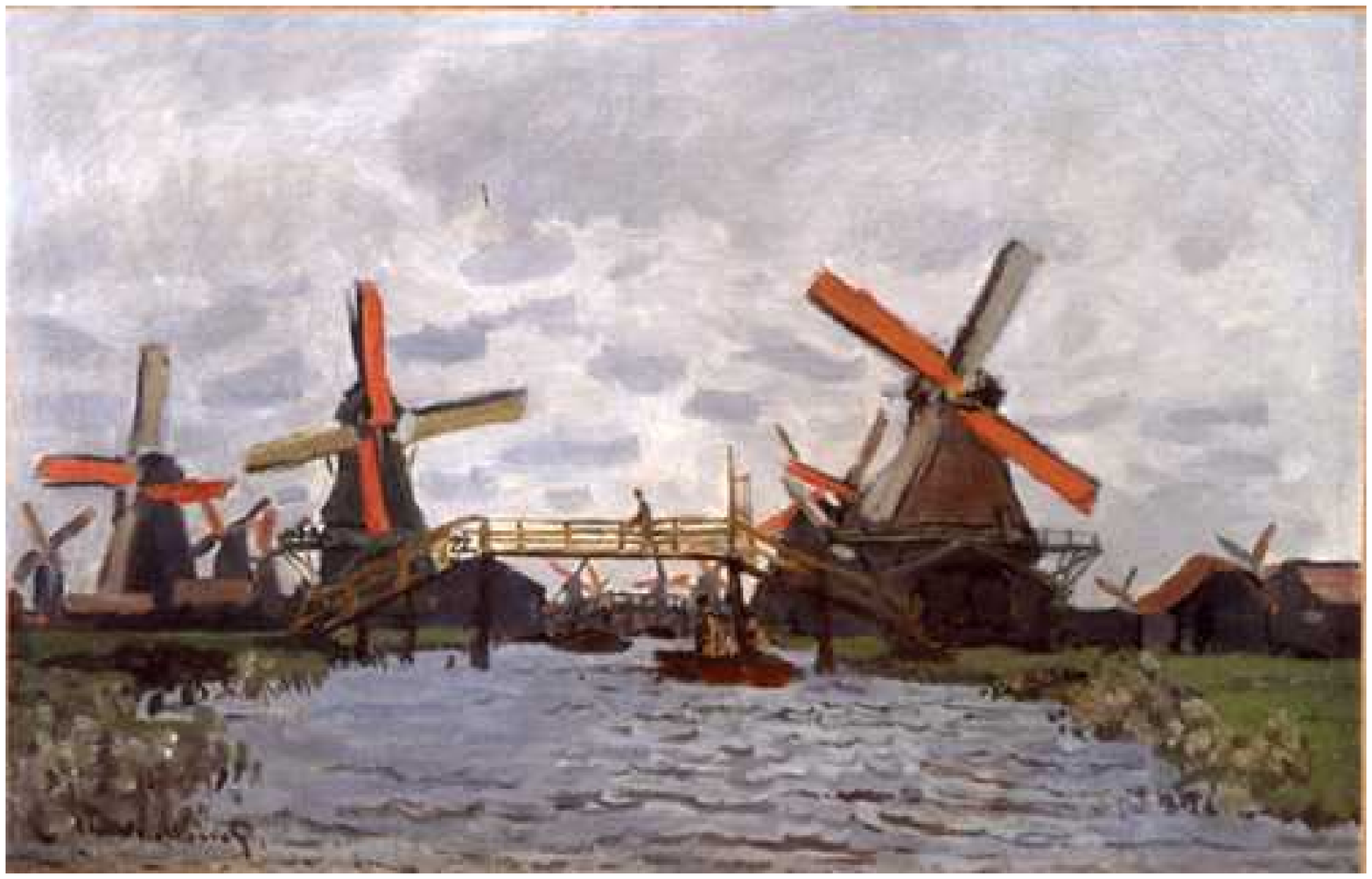}}
\subfigure[s206v]{
\label{Fig.sub5}
\includegraphics[width=0.20\textwidth]{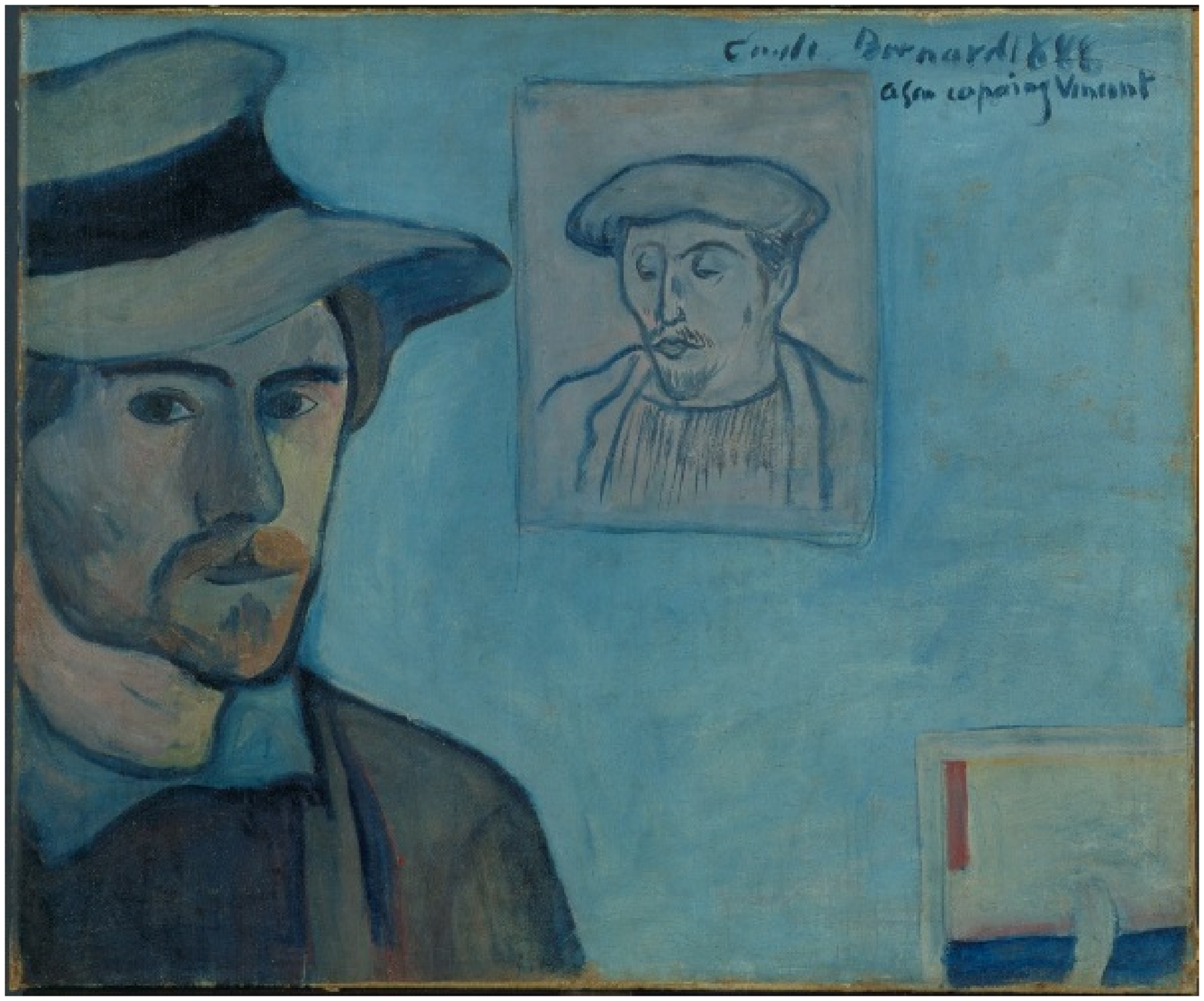}}
\subfigure[s225v]{
\label{Fig.sub6}
\includegraphics[width=0.20\textwidth]{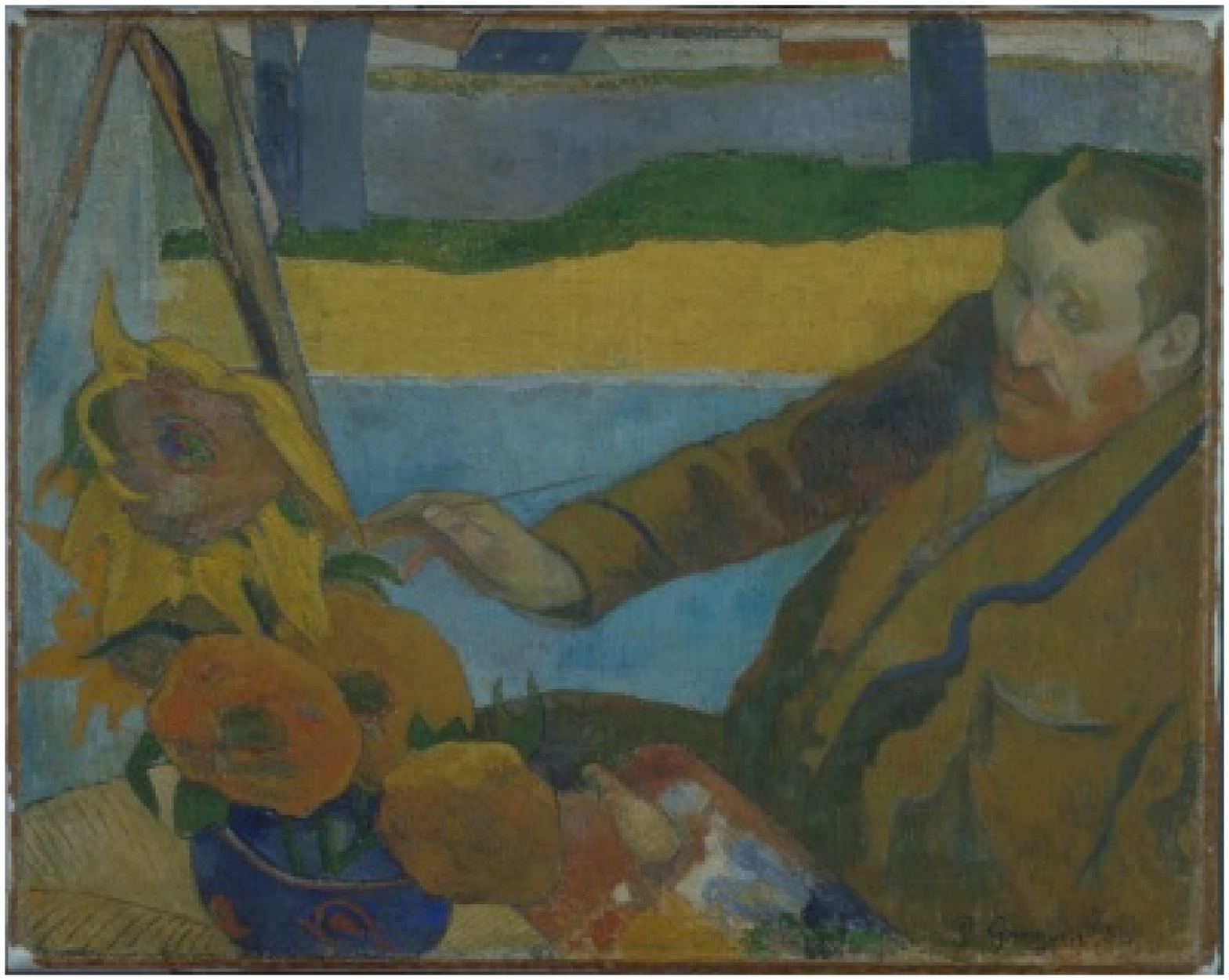}}
\subfigure[s205v]{
\label{Fig.sub7}
\includegraphics[width=0.20\textwidth]{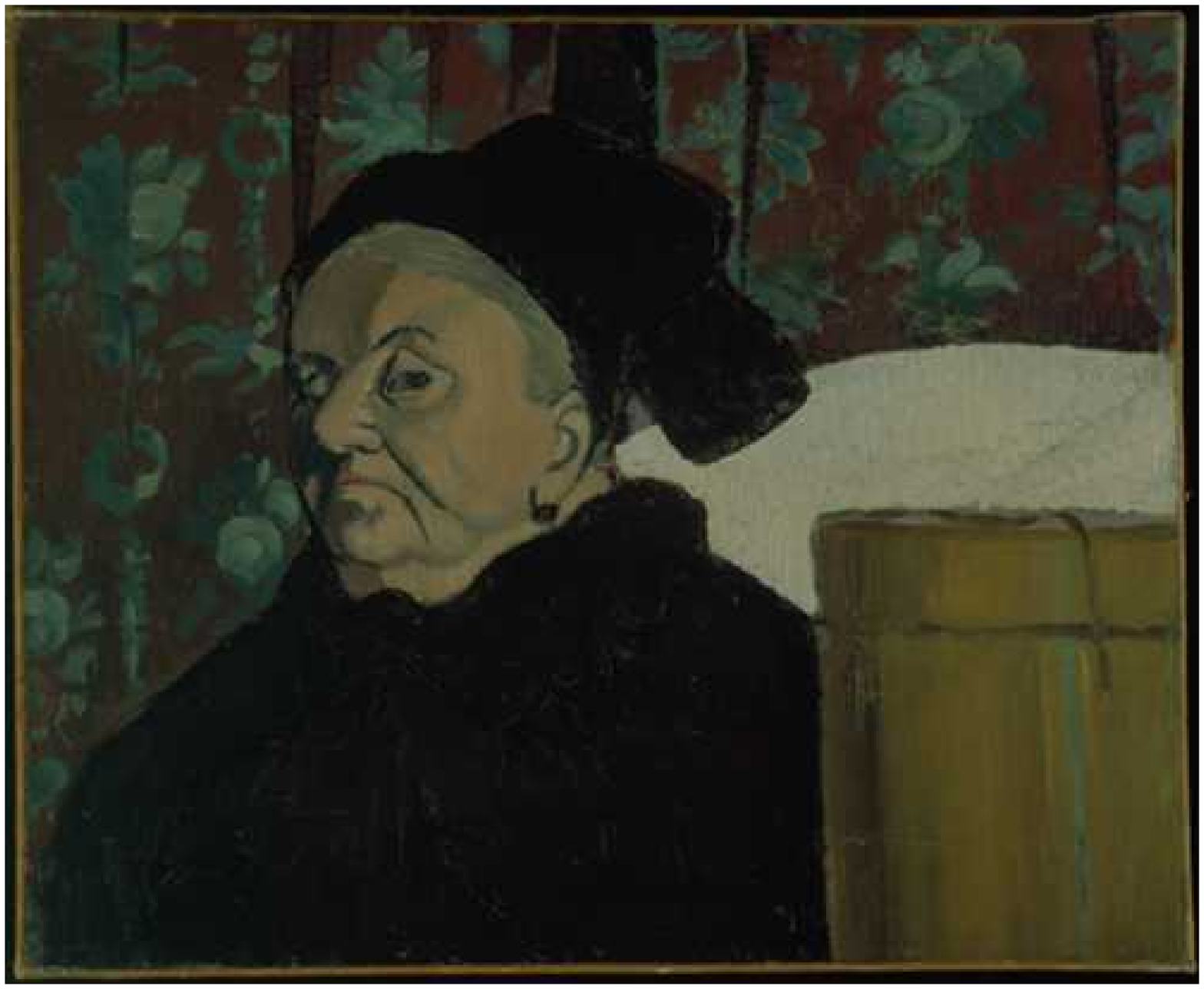}}
\caption{The seven forgeries detected as van Gogh's painting by our method.}
\label{fig:nvg}
\end{center}
\end{figure}

In Figures \ref{fig:vg} and \ref{fig:nvg}, we give
the misclassified paintings together with their ID numbers.
In particular, there are 3 forgeries (f253a, f418, and f687)
which were once wrongly regarded as van Gogh's paintings
and are indeed highly similar in such a stylometric analysis,
see Table \ref{nvg-confused}.
They successfully cheat both experts and our method.
On the other hand, van Gogh's paintings f249,
f371 and f752
exhibit so many unusually diverse movements of brushstroke that
they look different from van Gogh's other paintings.

Comparing the results in Table \ref{result}, the TPR of our method
is quite high while its TNR is a bit low. In Section \ref{analysis}, we show that we can significantly improve the TNR while keeping the TPR by carefully selecting the features.

\subsection{Feature analysis}
\label{analysis}
Recall that in our method for each test painting a set of five features $\cS$
is selected, see Section \ref{fss}.
In order to identify the most discriminatory features which are useful in
accurate classification, we gather the features sets $\cS$ in all 79 tests
and count the frequency of each feature that
occurs in these 79 $\cS$'s. It turns out that only 11 out of the 54 features
occur in these 79 feature sets and they are listed  in {Table}  \ref{feature-12}.

\begin{table}[h]
\begin{center}
\begin{tabular}{c|c|c}
filter & statistics & frequency \\\hline
${\tau_3}$ & { mean} & 78\\\hline
${\tau_{16}}$ & {percentage of the tail entries} & 77 \\\hline
${\tau_0}$ & {percentage of the tail entries} & 77 \\\hline
${\tau_1}$ & {percentage of the tail entries} & 76\\ \hline
${\tau_8}$ & {percentage of the tail entries} & 75\\\hline
$\tau_3$ & percentage of the tail entries & 5 \\\hline
$\tau_1$ & mean & 3\\ \hline
$\tau_4$ & mean & 1 \\\hline
$\tau_{9}$ & mean  & 1\\\hline
$\tau_{17}$ & mean  & 1\\\hline
$\tau_{2}$ & mean  & 1
\end{tabular}
\vspace*{10pt}
 \caption{The eleven features (with their filter and statistics)
 and their frequencies of occurrence.}\label{feature-12}
\end{center}
\end{table}

From the table, we see that the first five features occurs with
an average of 96.96\% frequency (total 383 occurrences out of $5\times 79$).
Thus they are the most discriminatory features. Notice that the features
on standard derivation are not selected at all. Thus in hindsight, we could start our method
with only 2 statistics, i.e. the mean $\mu^{(i,j)}$ and the percentage $p^{(i,j)}$ (see
(\ref{fv})), and end up with the same result.

To test the discriminatory power of the top features in {Table}  \ref{feature-12},
we use the top four and five features to do the
classification under the leave-one-out cross-validation. They give the
accuracies of 86.08\% and 88.61\% (see Table \ref{result1}), respectively.
Indeed using the top five features, we can further identify
Figure \ref{fig:vg} (f) and (g) to be forgery, thus improving the TNR.
The accuracy of our method using only the top five features (i.e. 88.61\%) is better than the best ones (87.69\% for IP4AI1 and 85.0\% for IP4AI2) obtained by others so far or by us (86.08\%), see Table \ref{result}.

The success of this small set of five
features reflects a highly consistent style in van Gogh's
brushstroke movements, where many forgeries demonstrate a more diverse spread in these features. Our method also leads to an interpretable model: e.g. the first feature
in Table \ref{feature-12} is
related to filter $\tau_3$ (see (\ref{filters})) indicating that the $-\pi/4$ direction
is an important direction in discriminating van Gogh's paintings.

\begin{table}[h]
\begin{center}
\begin{tabular}{c|c|c|c}
   &TPR\ (TP)  & TNR\ (TN) & classification \\
 &                        &                 & accuracy (TP$+$TN) \\\hline
  Our method&  93.75\% (60) & 53.33\% (8) &  86.08\% (68) \\   \hline
4-feature classifier & 92.19\% (59)& 60.00\% (9)& 86.08\% (68) \\\hline
5-feature classifier & 93.75\% (60)& 66.67\% (10) & 88.61\% (70)
\end{tabular}
 \caption{Classification results on different sets of classifiers.}\label{result1}
\end{center}
\end{table}

\subsection{Method evaluation}
\label{evaluation}

In this section, we statistically evaluate our method and the top 5 features we selected
in Section \ref{analysis}.
Since we only have one dataset of 79 paintings, we generate
200 similar datasets by bootstrap sampling with replacement \cite[p.12]{efron1994introduction}.
More precisely, each of these 200 datasets are generated by randomly choosing 64 samples from
the 64 vG paintings with replacement and 15 samples from 15 nvG painting with
replacement. We will compute the accuracy of our method and also the accuracy
of the top 5 features
on these 200 datasets. Suppose the accuracy of the $i$-th dataset is $y_i$, $i=1,\cdots,200$.
The 95\% confidence interval we give below is defined to be $(y_{i_6}, y_{i_{195}})$ with $y_{i_1}\le\cdots\le y_{i_{200}}$, see \cite{buckland1984monte}.

 \subsubsection{Evaluation of our method}
\label{evaluation_method}
As mentioned above, there are 200 randomly chosen datasets for evaluation of our method,
 with each dataset having 79 samples and 54 features.
For each such dataset, the accuracy of our method is again tested by using the leave-one-out cross-validation described in Sections \ref{extraction}--\ref{classification}, where a painting in the dataset is used as a testing data and the remaining 78 samples are the training data to select the five important features. Then a classifier is trained for authentication of the left-out painting. This procedure is repeated
79 times until each painting in the 79 samples is tested once and a classification accuracy value is obtained from these 79 tests. Figure \ref{Fig.lable} (left) is the histogram of the 200 accuracy values for these 200 datasets. The mean, median and standard deviation are 83.73\%, 83.54\% and 0.0507 respectively.
 The 95\% confidence interval is (73.42\%, 92.41\%).

\subsubsection{Evaluation of the top five features}
\label{evaluation_five}
\begin{figure}[h]
\centering
\subfigure{\includegraphics[width=0.48\textwidth]{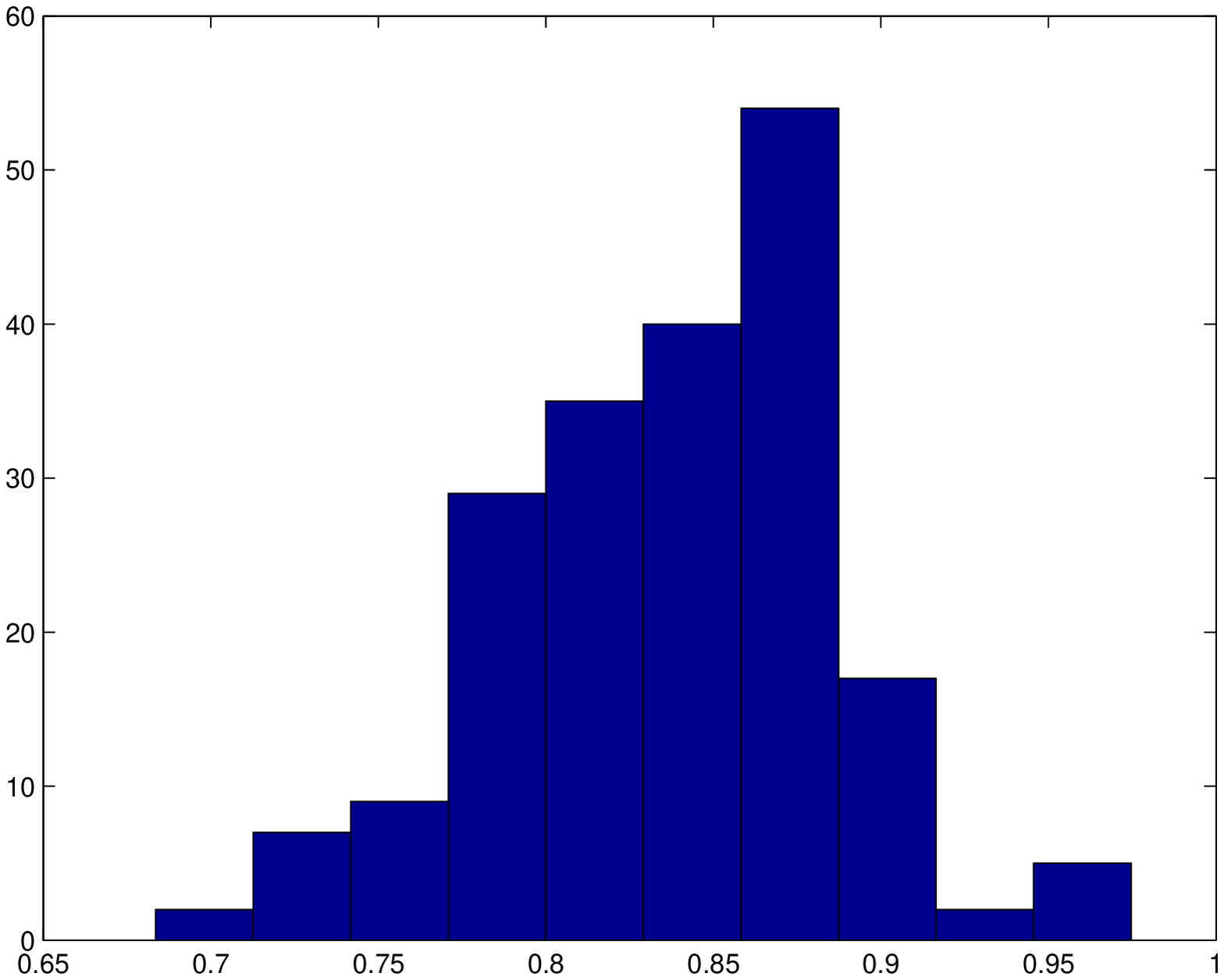}}
\subfigure{\includegraphics[width=0.48\textwidth]{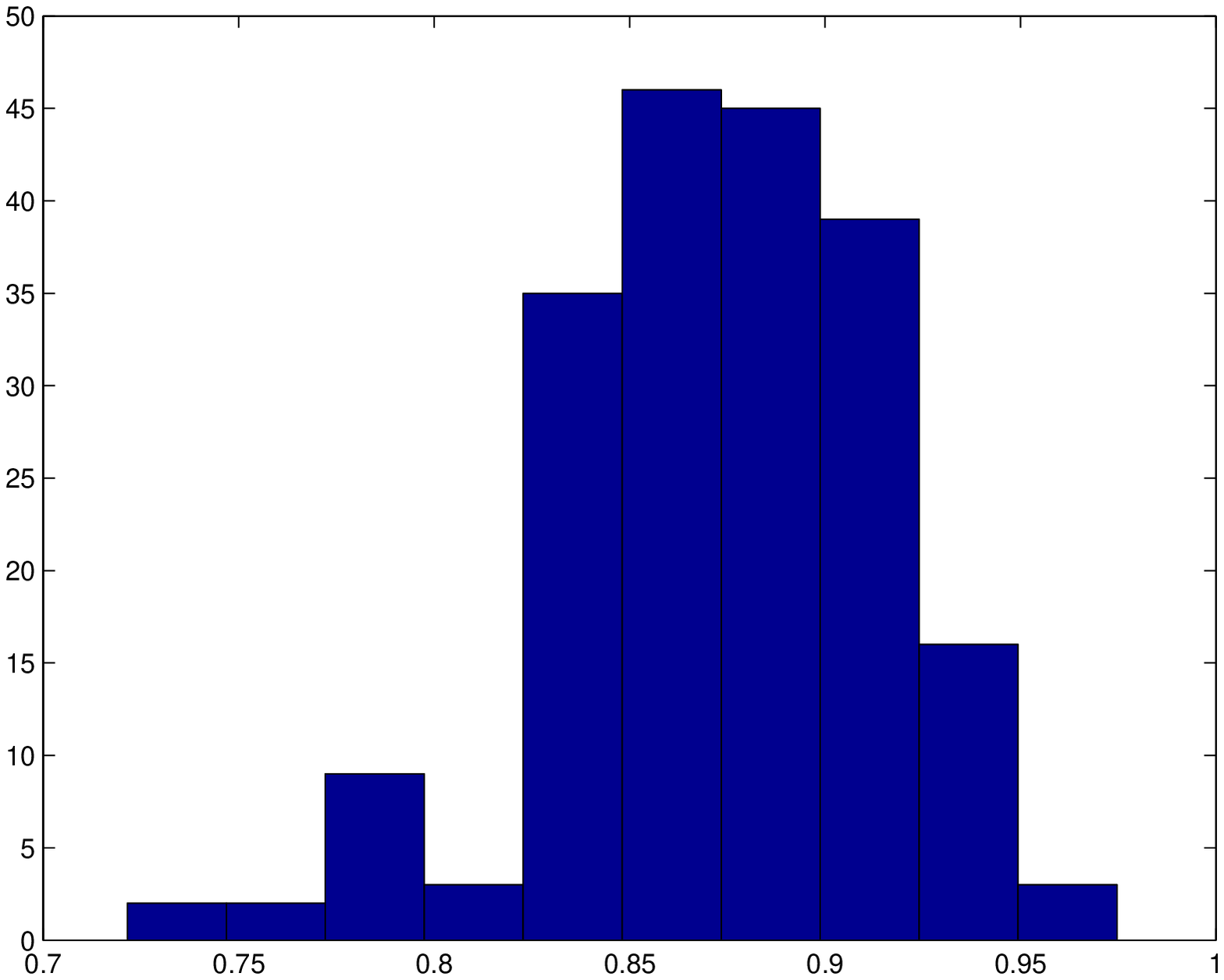}}
\vspace*{-7pt}
\caption{Histograms of the accuracy values for (i) our methods (left)
and (ii) for the top 5 features (right).}
\label{Fig.lable}
\end{figure}

Here we evaluate the top five features given in Table \ref{feature-12}. Similar to Section \ref{evaluation_method}, for each of the 200 datasets, the accuracy is tested under
the leave-one-out cross-validation, except that everything is done w.r.t the top five features only. Figure \ref{Fig.lable} (right) is the histogram of the 200 accuracy values. The mean, median and standard deviation are 87.77\%, 88.61\% and 0.0435 respectively. The 95\% confidence interval is (78.48\%, 94.94\%). Using only the best five out of the 54 features, the classification accuracy is quite good and  the results reflect the consistency of van Gogh's habitual brushstroke movements.

\section{Conclusion}
\label{conclusion}
We have proposed a geometric tight frame based visual stylometry
method to discriminate paintings by van Gogh from those
by imitators. The methodology consists of some simple statistics of the geometric tight frame coefficients,
as well as a boosting procedure for feature selection.
Our methodology has been tested on a data set of 79 paintings provided
by the van Gogh museum and Kr\"{o}ller-Muller museum.
The classification accuracy of our method is $86.08\%$.
The high classification accuracy shows that our features
are appropriate in identifying the authorship of van Gogh's paintings.
In particular, our method identifies five robust features
such that van Gogh's paintings show a higher degree of similarity in that
feature space  while forgeries exhibit a wider spread tendency as outliers.
The accuracy using these 5 features is 88.61\% which is the best one compared with the existing methods so far (see Tables \ref{result} and \ref{result1}).
 The success of this small set of features reflects the consistency of
van Gogh's habitual brushstroke movements.  From our results, we
 see that the ``statistical outliers" of certain tight frame coefficients are not noise, but important signals to distinguish van Gogh's painting from his contemporaries. Such ``outliers" and their tail distributions may due to the intrinsic creativity of
the maestro expressed through his brushstroke styles.
We hope these features may help art scholars to find new digital evidences
in van Gogh's art authentication. Our methodology can easily be
generalized to authenticate paintings for other artists and
that will be our future research directions.

\vspace{5mm}
\noindent
{\bf Acknowledgement:} We thank {Profs. Haixiang Lin and Eric Postma} for their helpful discussions
and providing us with the 79 paintings used in this paper.

\bibliographystyle{siam}
\bibliography{my}


\end{document}